# An Exploration of Clustering Algorithms for Customer Segmentation in the UK Retail Market

Jeen Mary John [1], Olamilekan Shobayo [1,*] and Bayode Ogunleye [2]

1 Department of Computing, Sheffield Hallam University, Sheffield S1 2NU, UK; jeenmary.m.john@student.shu.ac.uk
2 Department of Computing & Mathematics, University of Brighton, Brighton BN2 4GJ, UK; b.ogunleye@brighton.ac.uk
* Correspondence: o.shobayo@shu.ac.uk

**Abstract:** Recently, peoples' awareness of online purchases has significantly risen. This has given rise to online retail platforms and the need for a better understanding of customer purchasing behaviour. Retail companies are pressed with the need to deal with a high volume of customer purchases, which requires sophisticated approaches to perform more accurate and efficient customer segmentation. Customer segmentation is a marketing analytical tool that aids customer-centric service and thus enhances profitability. In this paper, we aim to develop a customer segmentation model to improve decision-making processes in the retail market industry. To achieve this, we employed a UK-based online retail dataset obtained from the UCI machine learning repository. The retail dataset consists of 541,909 customer records and eight features. Our study adopted the RFM (recency, frequency, and monetary) framework to quantify customer values. Thereafter, we compared several state-of-the-art (SOTA) clustering algorithms, namely, K-means clustering, the Gaussian mixture model (GMM), density-based spatial clustering of applications with noise (DBSCAN), agglomerative clustering, and balanced iterative reducing and clustering using hierarchies (BIRCH). The results showed the GMM outperformed other approaches, with a Silhouette Score of 0.80.

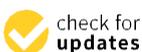

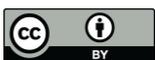

## 1. Introduction

The UK's retail market has been known to create a sizable amount of employment and sales. In the era of Big Data, decision making based on data is becoming increasingly important for organisations, especially the retail industry. Big Data Analytics (BDA) is used to analyse massive datasets and identify patterns in data, which can be used to make informed decisions [1]. Segmentation is a well-established marketing strategy that involves dividing customers into distinct groups, known as market segments, and focusing marketing efforts towards the most favourable segment [2]. Gaining insight into consumer behaviour and the decision-making process is crucial in formulating efficient strategies to deliver a customer-centric service and enhance profitability. Business Intelligence (BI) and Big Data Analytics (BDA) are employed in the customer segmentation framework to discern the prospective customer profile and market both new and established products. However, there is a lack of contextual framework behind practice development, specifically regarding the influence on decision marketing strategic issues.

Online customer data could be used to generate personas that represent distinct groups of individuals [3,4]. Business Intelligence (BI) and Big Data Analytics (BDA) could also be leveraged for the creation of a framework aimed at discerning the prospective customer profile for the purpose of marketing both new and established products [4].

This study aims to develop a customer segmentation model to improve decision-making processes in the retail market industry. To this end, we conduct an experimental comparison of various unsupervised machine-learning-based clustering algorithms for retail customer segmentation. Hicham and Karim [5] and Turkmen [6] compared some clustering approaches for customer segmentation. Their studies achieved Silhouette Scores of 0.72 and 0.6, respectively. These studies are similar to ours, but differ in terms of methodology and results. Thus, our contributions can be summarised as follows. We performed a comprehensive analysis of several algorithms for retail customer segmentation and discuss them from both a scientific and business perspective. We demonstrated the use of principal component analysis (PCA) and the Gaussian mixture model (GMM) and achieved a more accurate model and interpretable results. Subsequent sections present the background knowledge to this study (Section 2), the methodology (Section 3), and the results (Section 4), and Section 5 will provide conclusions and recommendations.

## 2. Related Work

Retailers in the UK must compete on pricing to stay afloat during the present economic crisis. To assist the economy, client loyalty research is essential. According to a study conducted in the UK, consumers are more devoted to specific retailers and brands when purchasing is convenient. The cost of the goods is not as significant as service operations and customer pleasure. To successfully control consumer purchasing behaviour and boost retail sales, retail companies must strengthen service operations skills [7]. By allowing personalised shopping experiences, predicting patterns, and taking wise choices based on market information, BD has transformed the retail industry [8].

During the era of Big Data (BD), decision making driven by data is common, irrespective of the scale of the business or the industry, who will be able to tackle complex business problems by taking actions based on data-driven insights [9]. According to Oussous et al. [10], BD, in contrast to traditional data, refers to vast, expanding data collections that encompass many different diverse forms, including un-structured, semi-structured, and structured data.

A case study was conducted by Jin and Kim on a typical courier company's sorting and logistics processing [11]. According to the study, BD can enhance business efficiency by transforming raw data into valuable information. Identifying the type and scope of data is crucial for achieving sustainable growth and competitiveness. Integrated use of BI, BD, and BDA in management decision support systems may aid businesses in achieving time- and cost-effectiveness. This case study provides useful insights for future company plans to reduce trial-and-error iterations. The authors of [12] developed a generic framework for organisational excellence by integrating Baldrige and BI frameworks. They formed respective matrices and adapted the Baldrige and BI frameworks, incorporating knowledge management and BI frameworks with specific key performance indicator (KPI) parameters. The framework integrates KPIs from the Baldrige framework, customer management, workforce engagement, knowledge management, operations focus, strategic planning, and academic accreditation. The dashboard is designed as an infographic mechanism, incorporating business, content, analytics, and continuous intelligence.

A K-means clustering (KC)-based consumer segmentation model was proposed in [13] and implemented using RStudio. The CSV-formatted dataset was analysed using various R functions. Visualisations were developed to understand client demographics. To identify the clusters, the elbow technique, average silhouette method, and gap statistics were used.

To examine the use of BDA in supply chain demand forecasting, the authors of [14] employed neural networks (NNs) and a regression analysis approach. The research findings are based on closed-loop supply chains (CLSCs) and provide suggestions for future research.

Customer segmentation is crucial in marketing as it helps organisations understand and meet their consumers' demands by breaking down an intended market into groups based on shared traits. Machine-learning-based customer segmentation models, such as k-means clustering, density-based spatial clustering of applications with noise (DBSCAN), and balanced iterative reducing and clustering using hierarchies (BIRCH), have presented potential insights in analysing customer data for effective decision making in the mar-keting sector. Ushakovato and Mikhaylov [15] developed a Gaussian-based framework for analysing smart meter data for predicting tasks. The experimental study analysed and compared different clustering techniques for predicting time series data and cluster assignments without additional customer information. Fontanini and Abreu [16] used the BIRCH algorithm to determine common load forms in neighbourhoods. The global clustering measure was developed by solving issues generated during optimisation of load forms. Lorbeer et al. [17] found that BIRCH requires the maximum number of clusters for a better clustering quality performance. This technique extracts load forms from large databases, clusters high, moderate, and minimal load types using cost functions, and determines the right number of clusters for global grouping. It is suitable for urban-scale loading assessments and real-time online education. Hicham and Karim [5] proposed a clustering ensemble method which consists of DBSCAN, k-means, MiniBatch k-means, and the mean shift algorithm for customer segmentation. They applied their clustering ensemble method to 35,000 records and achieved a Silhouette Score of 0.72. The authors of [18] developed a customer segmentation model (RFM+B) using indicators of RFM and balance (B) to enhance marketing decisions. This model classifies client savings using transactional patterns and current balances using the RFM and B qualities. The model achieved an accuracy of 77.85% using the K-means clustering technique. Hossain [19] segmented customer data based on spending patterns through k-means clustering and the DBSCAN algorithm, determining clients with out-of-the-ordinary spending patterns. This study suggested incorporating neural-network-based clustering techniques for customer segmentation on large datasets. Punhani et al. [20] applied k-means clustering to 25,000 online customer records obtained from Kaggle repository. They used the Davies–Bouldin Index (BDI) to evaluate their optimal number of K and thus segmented the customer records into four segments. Turkmen [6] compared DBSCAN, k-means clustering, agglomerative clustering, and the RFM framework with 541,909 retail customer datasets. They employed random forest for selection of variables fitted into the algorithms. They showed k-means clustering achieved the best result with a Silhouette Score of 0.6. In summary, our literature review findings suggest there is limited comparative studies of sophisticated unsupervised learning approaches for customer segmentation in the retail marketing context. Thus, subsequent sections will present our methodology and the results.

## 3. Methodology

This section presents the methods adopted in this study. The concept is to segment customers based on their attributes (purchase behaviour). Thus, unsupervised machine learning clustering techniques are considered appropriate for this purpose. We employ sophisticated clustering algorithms, namely, k-means clustering, the Gaussian mixture model (GMM), density-based spatial clustering of applications with noise (DBSCAN), agglomerative clustering, and balanced iterative reducing and clustering using hierarchies (BIRCH). Beforehand, we employed the recency frequency monetary (RFM) model. We will thus provide a comprehensive analysis of the ML-based clustering models for customer segmentation. In the subsequent section, we will discuss the concepts of these algorithms.

### 3.1. RFM

The recency, frequency, monetary (RFM) model is a valuable tool for client segmentation, focusing on three key factors, namely, recent purchases, transaction frequency, and spending amount [19]. It helps companies identify distinct client segments and develop customised marketing strategies. The RFM model's simplicity and efficiency enable companies to allocate resources more effectively, focusing on high-value consumers and fostering customer loyalty. To enhance segmentation and gain deeper insights, the model can be supplemented with new dimensions or complementary methods like clustering algorithms. This approach has gained significant attention in shaping specialised marketing strategies [6]. Figure 1 below provides a simple illustration of the RFM model.

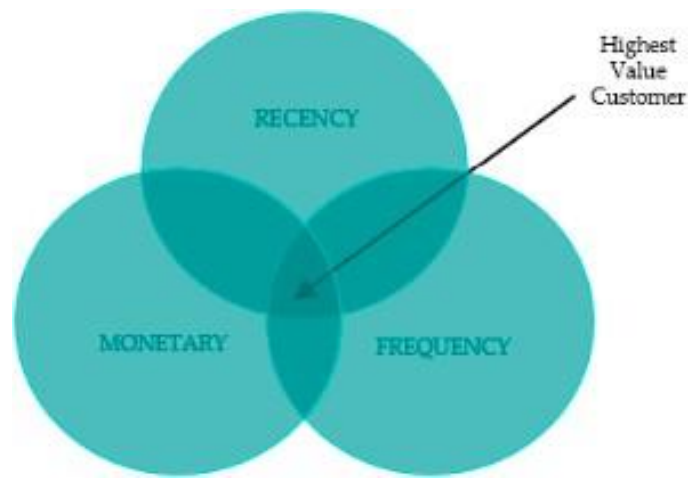

**Figure 1.** RFM model.

*3.2. Principal Component Analysis*

Principal component analysis (PCA) is a statistical technique that leverages the covariance component of a multivariate data, although they are different variants of the algorithm, for example, that use multi-layer perceptron to compute the linear auto-associative properties [21]. The principal components of multivariate data are computed and the subspace in one-dimension, where the data have the highest variance, is the first principal component. The subsequent principal components, i.e., second, third nth principal components, follow in the direction of the highest variance in space that is orthogonal to the previous principal component. The PCA is calculated by the eigenvalue of the input multivariate data by decomposition of the covariance matrix [22].

PCA works on the standardised range of values to allow for equal contribution for analysis. To obtain the PCA, the covariance matrix is first computed.

Given two-dimensional variables $x_1$ and $x_2$, the covariance can be computed as thus:

$$Cov(x_1, x_2) = \frac{\sum_{i=1}^{n}(x_{1i} - \overline{x_1})(x_{2i} - \overline{x_2})}{n-1} \quad (1)$$

For multivariate data, the covariant matrix can be developed. Let us consider $n$-dimensional data that can be obtained by:

$$\frac{n!}{(n-1)! \times 2} \quad (2)$$

Additionally, the covariance matrix for the set of data is given by:

$$C^{n \times n} = (c_{1i,j}, \; c_{1i,j} = cov \; Dim_i, \; Dim_j$$

where $C^{n \times n}$ is a matrix with $n$ rows and $n$ columns. The covariance matrix of a three-dimensional variable $x_1, x_2, x_3$ is thus represented as

$$C = \begin{bmatrix} cov(x_1, x_1) & cov(x_1, x_2) & cov(x_1, x_3) \\ cov(x_2, x_1) & cov(x_2, x_2) & cov(x_2, x_3) \\ cov(x_3, x_1) & cov(x_3, x_2) & cov(x_3, x_3) \end{bmatrix} \quad (3)$$

The corresponding eigenvalues $\lambda$ and eigenvectors $X$ of the covariance matrix can be expressed using the formula:

$$|A - \lambda I| = 0 \quad (4)$$

$$AX = \lambda X \quad (5)$$

where $A$ is an $n \times n$ square matrix and $I$ is an identity matrix of the same shape as $A$.

*3.3. Clustering Techniques*

Cluster analysis is an iterative process that divides data into clusters based on similar attributes [23]. Techniques include density-based, partition-based, and hierarchical-based algorithms. Partition-based clustering decomposes data points into distinct K groups. The approach requires the number of clusters (K) to be pre-defined. Hierarchical clustering gradually creates clusters. This can be by adding up nearest data points from the bottom to the top (agglomerative) or by splitting data points from the top (divisive). In this case, the number of clusters required or needed is not pre-defined. In contrast, density-based clustering groups closely packed data points in space. This technique examines irregular shapes and is less sensitive to noise samples.

3.3.1. K-Means Algorithm

The K-means algorithm enables the identification of homogenous groups within a consumer base [23]. It splits a dataset into discrete clusters based on similarity, enabling personalised marketing activities that cater to individual tastes and requirements [24]. The algorithm selects initial centroids, updates centroids based on the mean of the allocated points, and iteratively assigns data points to the closest cluster centroid [23]. The objective is to find K cluster centroids that minimise the sum of squared distances between data points and their respective cluster centroids. The within-cluster sum of squares (WCSS) is determined by dividing the sum of squared distances by the number of clusters.

The following is how the WCSS is determined:

$$WCSS = \sum_{i=1}^{N} \sum_{k=1}^{K} r_{ik} \cdot ||x_i - \mu_k||^2 \tag{6}$$

where:

$N$: number of data points.
$K$: number of clusters.
$x_i$: $i$th data point.
$\mu_k$: $k$th cluster centroid.
$r_{ik}$: indicator variable that is returned as 1 if data point $x_i$ belongs to cluster $k$ and 0 otherwise.

The K-means algorithm aids in customer profiling in consumer segmentation, helping firms identify segments with distinctive behavioural patterns, purchase frequencies, and preferences.

3.3.2. Gaussian Mixture Model (GMM) Algorithm

The Gaussian mixture model (GMM) is a powerful method for identifying subtle patterns in consumer datasets for customer segmentation. It captures the probabilistic distribution of data points, revealing hidden structures and linkages within customer behaviours. The GMM assumes a Gaussian distribution of data points within a cluster, finding hidden patterns that simpler algorithms might miss [25]. Its adaptability allows it to find clusters of various shapes, sizes, and densities, making it ideal for capturing intricate customer segmentation scenarios. For a dataset with $N$ data points and $K$ clusters, the GMM simplifies identifying client categories with complex behaviours, preferences, and interactions, enabling firms to adjust marketing campaigns to meet each segment's needs.

The GMM is defined as follows for a dataset with $N$ data points and $K$ clusters:

$$P(x) = \sum_{k=1}^{K} \pi_k \cdot N(x|\mu_k, \varepsilon_k) \tag{7}$$

where:

$\pi_k$: proportion of data points in each cluster.
$\mu_k$: mean vectors for each distribution.
$\varepsilon_k$: matrices that represent the shape and orientation of each distribution.

### 3.3.3. DBSCAN Algorithm

The density-based spatial clustering of applications with noise (DBSCAN) algorithm is a useful tool for identifying complex structures in consumer data for customer segmentation. It is particularly effective in recognising clusters of various shapes and densities, particularly in datasets with noise and irregularities [5]. DBSCAN groups data points based on their proximity and density, identifying core points with a minimum number of nearby points. It can identify both dense and sparse clusters, providing a comprehensive understanding of client behaviour trends. The reachability distance from point p to point q is calculated using the maximum of the core distance and Euclidean distance.

The maximum of the core distance of data point $p$ and the Euclidean distance $p$ and $q$ is the reachability distance from point $p$ to point $q$.

$$reach\_dist(p, q) = \max(core\_dist(p), ||p - q||) \tag{8}$$

### 3.3.4. BIRCH Algorithm

The balanced iterative reducing and clustering utilising hierarchies (BIRCH) algorithm is a powerful tool for handling large datasets quickly and maintaining hierarchical linkages in customer segmentation [17]. It uses a two-phase strategy, organising data points into subclusters and creating a Cluster Feature Tree (CFT) in the first phase. The second phase merges smaller clusters into larger ones, providing insights into broad and specific trends and scalable segmentation.

The formula for the distance between the two clusters $C_1$ and $C_2$ is:

$$D(C_1, C_2) = \sqrt{\frac{N_1 N_2}{N_1 + N_2} \cdot \frac{LSUM_1}{N_1} - \frac{LSUM_2}{N_2}} \tag{9}$$

where:

$N$: number of data points in the cluster.
$LSUM$: sum of data points in the cluster.

### 3.3.5. Agglomerative Algorithm

Agglomerative clustering is a popular hierarchical clustering method used in customer segmentation research. It treats each data point as an independent cluster and merges the nearest clusters iteratively until all data points are contained in a single cluster or a preset number of clusters are formed [26]. The distance between two clusters is computed using the algorithm's linking criteria, which include single linkage, complete linkage, average linkage, and Ward's linkage. Ward's linkage minimises the variance increase following cluster fusion, while the average linkage computes the average distance between all points.

The equations for these linkage approaches are as follows, given two clusters A $A$ and $B$ and $a$ distance metric $d(A, B)$ that gauges the distance between the clusters:

$$d_{single}(A, B) = \min_{a \in A, b \in B} d(a, b) \tag{10}$$

$$d_{complete}(A, B) = \max_{a \in A, b \in B} d(a, b) \tag{11}$$

$$d_{average}(A, B) = \sum_{a \in A} \sum_{b \in B} d(a, b) \tag{12}$$

*3.4. Dataset*

We used a retail dataset obtained from the UCI machine learning repository for our study. The dataset is publicly accessible at the UCI Machine Learning Repository: Online Retail Data Set (https://archive.ics.uci.edu/dataset/352/online+retail) (accessed on 16 May 2023). We chose the dataset due to its volume and popularity in proposing algorithms for retail customer segmentation [6]. The dataset contains 541,909 entries and eight different variables detailed below.

- InvoiceNo: Invoice number. A nominal, six-digit integral number uniquely assigned to each transaction. If this code starts with the letter 'c', it indicates a cancellation.
- StockCode: Product (item) code. A nominal, five-digit integral number uniquely assigned to each distinct product.
- Description: Product (item) name. Nominal.
- Quantity: The quantities of each product (item) per transaction. Numeric.
- InvoiceDate: Invoice date and time. Numeric, the day and time at which each transaction was generated.
- UnitPrice: Unit price. Numeric. Product price per unit in sterling.
- CustomerID: Customer number. A nominal, five-digit integral number uniquely assigned to each customer.
- Country: Country name. Nominal. The name of the country where each customer resides.

## 3.5. Experimental Set Up

Python is a robust tool for implementing advanced customer segmentation techniques due to its versatility, extensive libraries, and user-friendly syntax, making it an ideal choice for extracting valuable insights from complex datasets. Different python libraries were utilised in this research. Pandas was used for statistical analysis, NumPy for numerical calculations, StatsModels for data analysis, Matplotlib for charting, Scikit-Learn for ML algorithms and metrics evaluation, Seaborn for data visualisation, and Deap for evolutionary algorithms. Specialised imports include tools from the Deap library, modules like RandomizedSearchCV, LabelEncoder, MinMaxScaler, StandardScaler, PCA, KMeans, GaussianMixture, Silhouette_score, DBSCAN, and make_blobs, and various metrics. The warning module was imported to optimise code performance. The experimental setup is shown in Figure 2 below.

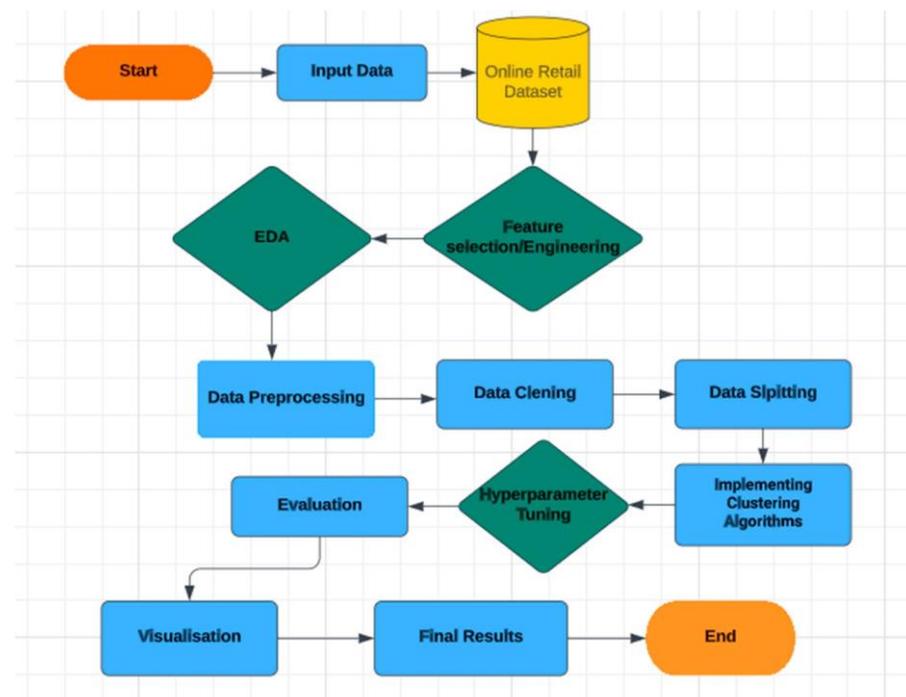

**Figure 2.** Experimental setup.

## 3.6. Data Pre-Processing

Data pre-processing is an essential step in ensuring a correct, consistent, and unbiased analysis. During the pre-processing stage, the null values were identified, specifically in the 'Description' and 'CustomerID' fields. The modified dimensions of the dataset are 406,829 rows and eight columns, indicating successful removal of blank "CustomerID" entries. Furthermore, incorrect values were removed from the dataset, with 8905 entries having negative values in the 'quantity' field. In addition, we eliminated duplicate records, removing rows with identical values in all columns. After pre-processing, we had a total of 397,924 records with eight columns. A sample data representation of the features is provided in Table 1 below.

**Table 1.** Important features in the retail dataset.

| Invoice No | Invoice Counts | Invoice Price (GBP) |
|---|---|---|
| 536365 | 7 | 139.122 |
| 536366 | 2 | 22.20 |
| 536367 | 12 | 278.73 |
| 536368 | 4 | 70.05 |
| 536369 | 1 | 17.85 |

*3.7. Model Evaluation*

The Silhouette Score is a metric used to assess the quality of clusters produced by various clustering techniques like k-means, DBSCAN, and Gaussian mixture models [6,27]. It measures the separation and compactness of the clusters, quantifying the cohesion of data points within the same cluster and separation between distinct clusters. Higher scores indicate better segmentation, aiding in selecting the best algorithm. The Silhouette Score is explained mathematically below:

$$s(i) = \frac{b(i) - a(i)}{\max(a(i), b(i))} \tag{13}$$

where:

$s(i)$ is the Silhouette Score for the data point.
$a(i)$ is the average distance between i and other data points in the same cluster.
$b(i)$ is the smallest average distance between i and data points in different clusters.

The Silhouette Score for the entire dataset is the average of the Silhouette Scores of all data points.

**4. Result**

This section presents our results for the RFM customer analysis and an experimental comparison of the unsupervised machine learning algorithms employed.

*4.1. RFM Results*

The RFM analysis includes recency, frequency, monetary value, and ranking customer scores based on the RFM scores. Customer value is segmented into top, high-value, medium-value, low-value, and lost customers based on their scores. Customer segments were annotated using nested NP. The input dataset was split into training and testing datasets at 80% and 20%, respectively. Each row displayed a unique customer and customer segment along with their score.

The process calculates the recency, frequency, and monetary values for each customer, determining their latest invoice date. Frequency is determined by counting distinct invoice dates for each customer. The total invoice price is represented as monetary. The first five records shown in Table 2 represent distinct customers with corresponding RFM values.

**Table 2.** RFM components.

| Customer ID | Recency | Frequency | Monetary (GBP) |
|---|---|---|---|
| 12346 | 325 | 1 | 77,183.60 |
| 12347 | 1 | 182 | 4310.00 |
| 12348 | 74 | 31 | 1797.24 |
| 12349 | 18 | 73 | 1757.55 |
| 12350 | 309 | 17 | 334.40 |

The study uses the rank method to rank customers based on their recency, frequency, and monetary worth. It assigns a rank to each customer based on their score, dividing each rank by the highest rank and multiplying it by 100. The first five entries in Table 3 represent distinct customers and their sum of recent, frequent, and monetary values. The RFM score was then used to rank customers' data. The RFM score was calculated using the following formula:

$$RFM\ Score = (Recency\ Score \times Recency\ Weight) + (Frequency\ Score \times Frequency\ Weight) + (Monetary\ Score \times Monetary\ Weight) \quad (14)$$

where:

*Recency Weight* reflects how important recency is in the analysis.
*Frequency Weight* reflects how important frequency is in the analysis.
*Monetary Weight* reflects how important monetary value is in the analysis.

**Table 3.** RFM score.

| Customer ID | RFM_Score |
|---|---|
| 12346 | 0.06 |
| 12347 | 4.48 |
| 12348 | 2.09 |
| 12349 | 3.41 |
| 12350 | 1.10 |

The customer segments were categorised based on their RFM score, with top customers having a score above 4.5, high-value customers having a score above 4, medium-value customers having a score above 3, low-value customers having a score above 1.6, and lost customers having a score below 1.6.

Table 4 below displays the top 10 records in the dataset, with each row representing a unique customer and the 'Customer_segment' column indicating their segment based on their score.

**Table 4.** Customer segmentation based on RFM score.

| Customer ID | RFM_Score | Customer_Segment |
|---|---|---|
| 12346 | 0.06 | Lost customer |
| 12347 | 4.48 | High-value customer |
| 12348 | 2.09 | Low-value customer |
| 12349 | 3.41 | Medium-value customer |
| 12350 | 1.10 | Lost customer |
| 12352 | 3.46 | Medium-value customer |
| 12353 | 0.03 | Lost customer |
| 12354 | 2.67 | Low-value customer |
| 12355 | 0.93 | Lost customer |
| 12356 | 3.11 | Medium-value customer |

Figure 3 displays the sizes and percentages of different client groups, with 31% of them being lost customers, 30% being low-value customers, 21% being medium-value customers, 10% being high-value customers, and 8% being top customers.

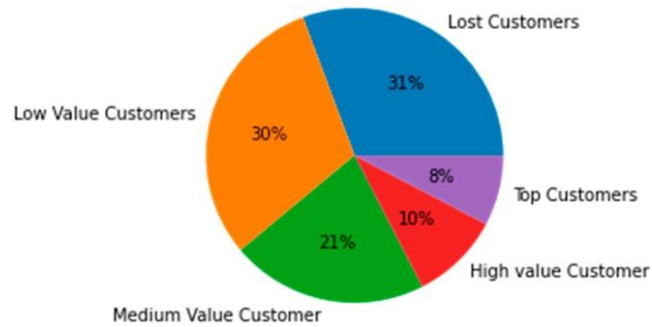

**Figure 3.** Visualisation of customer segments.

*4.2. Clustering Results*

Unsupervised machine learning algorithms were applied and evaluated using the Silhouette Score. Each clustering algorithm was evaluated by accessing the quality of clustering samples within the dataset.

4.2.1. K-Means Clustering

We employed the elbow approach to determine the optimal number of clusters (k). It estimates inertia, representing sample distances to the nearest cluster centre, for various cluster counts from 2 to 10. The inertia values were plotted against the number of clusters, as shown in Figure 3 below.

Figure 4 below illustrates the elbow method for visualising inertia, revealing that the optimal number of clusters is 3, as we are looking for a sharp edge like an elbow. We have scaled our dataset using Min–Max scaling for the dataset to conform to the same scale. This is a good practice for algorithms that rely on distance metrics for computation in order to avoid bias. The scaled data were then used to build our k-means clustering model. The result of the three clusters are visualised in Figure 5 below. Figure 5 displays a scatter plot of k-means, coloured by cluster labels assigned by the model. Inertia indicates the compactness of the clusters, while cluster centres reflect the centroid coordinates of each cluster.

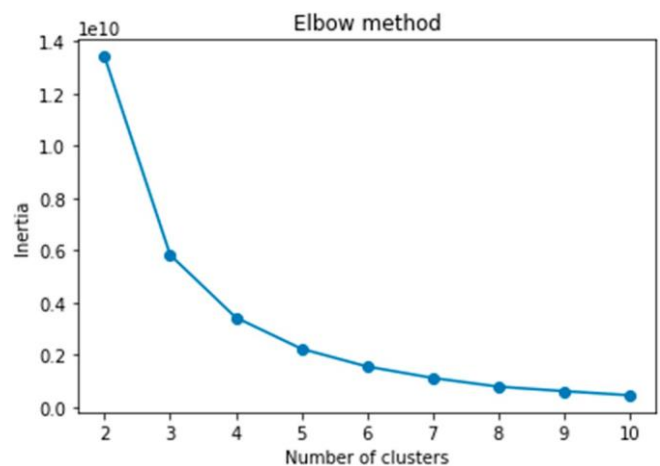

**Figure 4.** Elbow method.

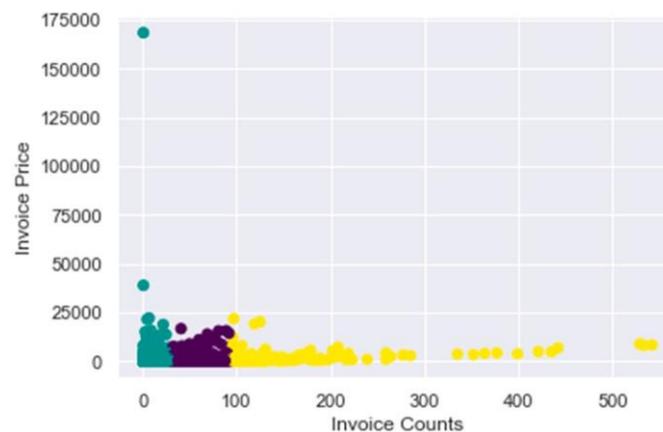

**Figure 5.** K-means algorithm visualisation.

4.2.2. Gaussian Mixture Model

Beforehand, we employed principal component analysis (PCA) for dimensionality reduction. The number of PCA components should not exceed the maximum possible number of components, which is determined by the minimum of the number of samples and features in the dataset. In this dataset, we have two features, so choosing two PCA components is a reasonable choice for visualisation purposes because it reduces the data to a two-dimensional space, making it easier to plot and interpret. Two principal components were found and thus these were fed into the GMM. Figure 6 below shows a visualisation of the GMM clustering results using P1 and P2 components. Each data point is assigned a colour based on its cluster.

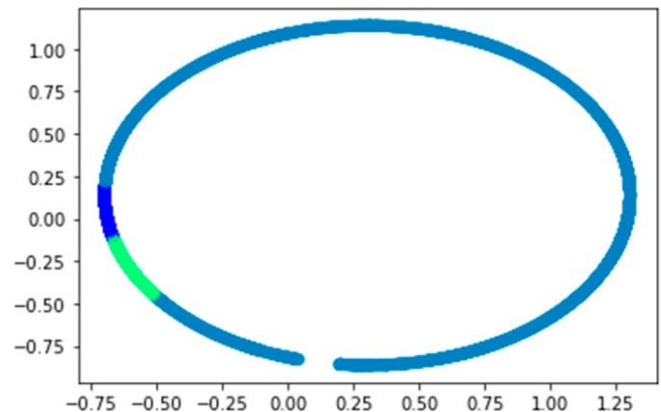

**Figure 6.** GMM result visualisation.

4.2.3. DBSCAN Clustering Algorithm

The DBSCAN clustering algorithm uses standardised data to compute cluster labels using the eps parameter to determine the maximum distance between samples in the same neighbourhood. We used a K-distance plot to determine the best epsilon (eps) value. While it was challenging to identify the knee point from the plot, we experimented with different epsilon values and found that the one yielding the highest accuracy was 0.3. The min_samples parameter specifies the minimum number of samples needed for dense regions. The algorithm assigns cluster labels to data points and identifies core samples with sufficient neighbours. The code segment calculates clusters by checking for noise points and subtracting one from the total number of unique labels to exclude noise clusters. Figure 7 below displays the DBSCAN clustering results, with markers representing data points and colours based on cluster labels. Noise points, represented by black, are not clustered and plotted as larger markers. The plot includes a title indicating the estimated number of clusters. There are three clusters: yellow, green, and red. Yellow clusters are between −2 and 0.5 on the x axis and −2 and 0.5 on the y axis. Green clusters are between (0, 0.5) and (2, 0.5), and red clusters are between 0 and 2. Visualising these results helps understand data distribution and separation.

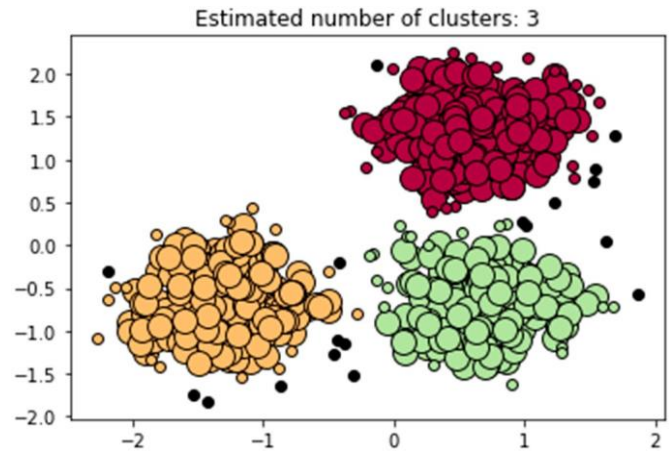

**Figure 7.** DBSCAN visualisation.

4.2.4. BIRCH Algorithm

The BIRCH clustering algorithm efficiently handles large datasets and is suitable for memory usage concerns. The threshold value determines the maximum diameter of subclusters in the hierarchical clustering structure. A lower threshold creates more fine-grained clusters, while a higher threshold leads to larger and coarser clusters. We tried different thresholds ranging from 0.01 to 1 and found that 0.01 yielded the highest performance based on the Silhouette Score. Three distinct groups have been identified in the dataset, as shown in Figure 8. The algorithm achieved a reasonable degree of separation based on the Silhouette Score, indicating a good clustering performance.

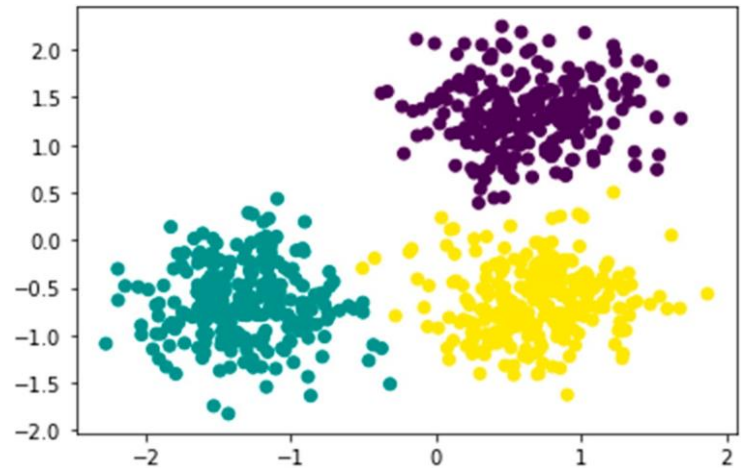

**Figure 8.** BIRCH algorithm visualisation.

4.2.5. Agglomerative Clustering

Figure 9 below shows the scatter plots of data points coloured according to their assigned cluster labels. This visualisation helps identify patterns, separations, and areas of similarities and dissimilarities. The "rainbow" colormap assigns different colours to each cluster, aiding in analysis and decision making.

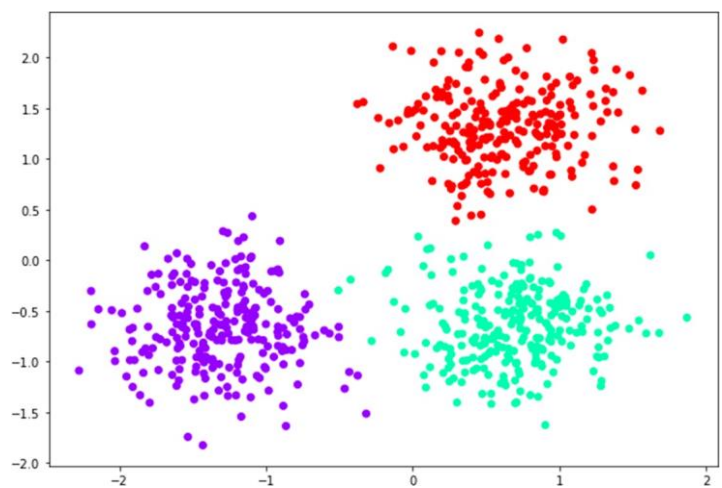

**Figure 9.** Agglomerative clustering visualisation.

*4.3. Model Performance Evaluation*

The Gaussian mixture model, k-means clustering, BIRCH, and agglomerative clustering created clusters in minimal computation times, making them better clustering algorithms in terms of complexity and the dataset used. Table 5 above shows the Silhouette Score for various clustering algorithms in our study compared to previous studies. We chose the Silhouette Score as our evaluation measure because it is commonly used in existing studies. This allowed us to compare the performance of our models with existing ones. The Silhouette Score of k-means is equal to 0.64, and those of the GMM model, the DBSCAN model, BIRCH, and the agglomerative algorithm are equal to 0.80, 0.626, 0.64, and 0.64, respectively. The Silhouette Score analysis revealed that our use of PCA and GMM achieved the best performance, with a Silhouette Score of 0.80. The GMM result showed cohesive datapoints and distinct clusters. We assessed the result visually and discovered that our approach produced elements that are closely packed. This implies that overlapping clusters or mislabelled data samples did not occur. This is because the Gaussian model captures the variance of the data, resulting in a better performance than other clustering models. BIRCH and DBSCAN consider the density and proximity of points rather than explicitly modelling distributions, which makes them less prone to issues related to high dimensionalities. GMM generally struggles with high-dimensional datasets. However, in our study, GMM benefitted from PCA (reduced dimension functionality). Several runs of the k-means algorithm were performed to check for improvements. The results varied by $\pm$ 0.06. This is due to the randomness in selecting the initial centroids. This is one of the limitations of k-means in the context of Big Data.

**Table 5.** Comparison of clustering algorithms for customer segmentation.

| Research Studies | Clustering Algorithm | Silhouette Score |
|---|---|---|
| Existing Studies | Mean shift [28] | 0.54 |
| | K-means [5] | 0.71 |
| | K-means [28] | 0.57 |
| | K-means [6] | 0.61 |
| | DBSCAN [5] | 0.72 |
| | Agglomerative [28] | 0.57 |
| Our study | K-means | 0.64 |
| | BIRCH | 0.64 |
| | DBSCAN | 0.62 |
| | Agglomerative | 0.64 |
| | Gaussian Mixture Model | 0.80 |

## 5. Conclusions

In this study, we aimed to develop a customer segmentation model to enhance decision-making processes in the retail market industry. To achieve this, we conducted an experimental comparative analysis of unsupervised machine learning clustering algorithms and thus showed that the use of principal component analysis (PCA) and the Gaussian mixture model (GMM) achieved a Silhouette Score of 0.8. The GMM provides useful insights into cluster formation. This paper also facilitates a data-driven decision-making approach through customer-data-based segmentation using machine learning techniques. This approach is beneficial to professionals performing customer segmentation, specifically in the retail market, as data are growing exponentially. One of the limitations of the RFM technique is its inability to reflect differences between the consumption and behaviour of customers. Less consideration is also given to business investments in customers. For future work, the use of sophisticated machine learning algorithms such as spectral clustering, which will allow for better analysis of customer behaviour, is proposed as they will help provide better insights into customer segmentation. Additionally, we aim to apply the kernel function to improve the linear separability of the GMM cluster output.

**Author Contributions:** Conceptualisation, J.M.J. and O.S.; Data curation, J.M.J.; Formal analysis, B.O.; Investigation, J.M.J. and O.S.; Methodology, J.M.J., O.S. and B.O.; Supervision, O.S.; Visualisation, O.S. and B.O.; Writing—original draft, J.M.J.; Writing—review and editing, O.S. and B.O. All authors have read and agreed to the published version of the manuscript.

**Funding:** This research received no external funding.

**Institutional Review Board Statement:** Not applicable.

**Informed Consent Statement:** Not applicable.

**Data Availability Statement:** The dataset used in this work is available at the UCI Machine Learning Repository: Online Retail Data Set and the code can be accessed publicly via this link: https://github.com/jeenmary55/Retail-Customer-Segmentaion (accessed on 2 September 2023).

**Conflicts of Interest:** The authors declare no conflict of interest.